\documentclass[conference]{ieeeconf}

\usepackage{graphicx}%, subfig}
\usepackage{amsmath, amsfonts, mathtools, bm}
\usepackage{algorithm, algpseudocode} % add in algorithm
\usepackage{booktabs} % multicolum
\usepackage{comment} % comment out multiple lines

\usepackage[none]{hyphenat} % no hypenation
\usepackage{listings} % Add code
\usepackage{xcolor}   % Add color to code

\usepackage{units}

\title{\bf Learning Stable and Energetically Economical Walking with RAMone }

\author{Audrow Nash$^*$, Yu-Ming Chen, Nils Smit-Anseeuw, Petr Zaytsev, and C. David Remy
    \\Robotics and Motion Laboratory (RAMlab), University of Michigan, Ann Arbor, MI
    \\$^*$audrow@umich.edu}

\begin{document}

\maketitle
\thispagestyle{empty}
\pagestyle{empty}

\begin{abstract}

    In this paper, we optimize over the control parameter space of our planar-bipedal robot, RAMone \cite{smit2017ramonehardware}, for stable and energetically economical walking at various speeds.
    We formulate this task as an episodic reinforcement learning problem and use Covariance Matrix Adaptation \cite{hansen1996adapting}. 
    The parameters we are interested in modifying include gains from our Hybrid Zero Dynamics style controller \cite{westervelt2003hybrid} and from RAMone's low-level motor controllers. 
    
\end{abstract}

\section{Introduction}

\begin{figure} [t]
        \centering
        \includegraphics[width=.23\textwidth]{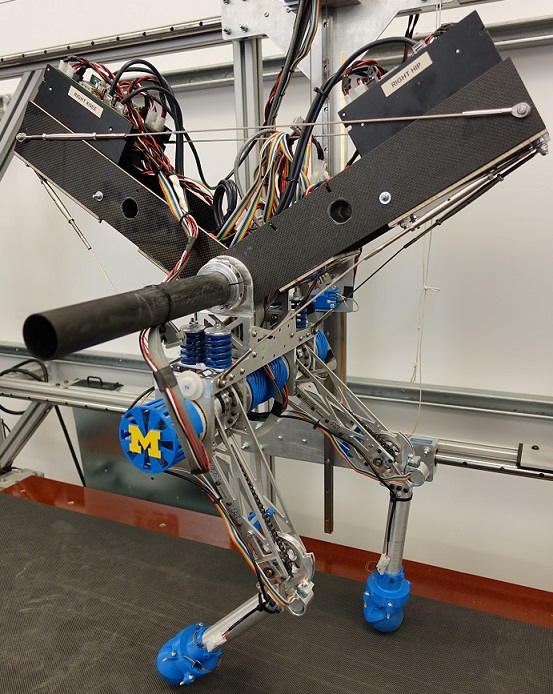}
        \caption{
        The robot RAMone \cite{smit2017ramonehardware} is a five-link biped with series elastic actuation at the knees and hips, and rolling contacts at the feet. 
        The robot is mounted on a planarizer system that restricts its motion to the sagittal plane \cite{green2016design}. 
        The RAMone hardware is based on the ScarlETH leg design \cite{hutter2011high}.
        }
        \label{Fig: RAMone Sim}
    \end{figure}

    Humans \cite{margaria1938sulla} and animals \cite{hoyt1981gait} use different gaits to locomote with energetic economy at different speeds. 
    In our previous work \cite{smit2017energetic}, we found that the same was true for a detailed model of the planar bipedal robot RAMone (Fig.~\ref{Fig: RAMone Sim}): walking was more economical for RAMone at low speeds, and running at high speeds. 
    However, it remains to be seen if these simulated results extend to RAMone in hardware. 
    
    To this end, we used numerical optimization to find energetically economical gaits for a model of RAMone at various speeds \cite{smit2017energetic}. 
    This optimization involved minimizing the energetic cost of transport (CoT), the electrical work needed to travel a unit distance. 
    The computed gaits described the optimal joint and motor trajectories as functions of time.
    
    These optimal trajectories are not sufficiently stable when run in an open-loop manner on RAMone. 
    One way to stabilize them, as we do here, is with a Hybrid Zero Dynamics (HZD) style controller, which synchronizes the controlled degrees of freedom to a phase variable \cite{westervelt2003hybrid}.
    
    An HZD style controller requires tuning a set of parameters, which at different speeds may have different values.
    Hand-tuning, although often used in practice, is time-consuming and may not lead to desired results.
    For example, we were unable to find parameters for stable walking at low speeds.
    Hand-tuning is likely to be even more of a challenge on hardware.
    
    In this work, we explore an automated method to optimize over our parameter space at various speeds in simulation. 
    We formulate this problem as an episodic reinforcement learning task. 
    Our plan is to use the resulting control parameters to achieve stable walking in hardware.
    
    %\subsection{Related Work}
    %    Other researchers have used reinforcement learning for legged locomotion \cite{kohl2004policy}. 
    %    Fankhauser et al. \cite{fankhauser2013reinforcement} uses a similar method to the one described in this abstract, in addition to using similar hardware, specifically the ScarlETH leg \cite{hutter2011high}. 
    %    They use reinforcement learning to optimize for CoT during a hopping task. 
    %    However, they formulate their optimization as a continuing reinforcement learning problem, rather than episodic, to improve learning on hardware.
        
\section{Method}

    \subsection{Parameter Space}
    
        For our walking controller, we follow a similar approach to \cite{westervelt2003hybrid}, which relates the optimal trajectories, discussed earlier, to a phase variable (here, this is horizontal displacement of the upper body from the stance foot). 
        This HZD controller has two gains to tune: the foot clearance gain $k_{fc}$ and the foot placement gain $k_{fp}$. 
        The foot clearance gain modifies the swing leg's knee angle to change the height of the swing foot trajectory; the foot placement gain modifies stance leg's hip angle to control the next stepping location. We also tune two low-level gains of the system: the proportional error tracking gains $k_{hip}$ and $k_{knee}$ of the hip and knee motors. 
        Lower gain values result in a more compliant controller; whereas higher gains result in a stiffer controller. 
        These four gains 
            $$K = [k_{fc}, k_{fp}, k_{hip}, k_{knee}]$$ 
        make up the parameter space for our optimization.
    
    \subsection{Optimization}
    
        To optimize over our parameter space, we use Covariance Matrix Adaptation, or CMA \cite{hansen1996adapting}. 
        CMA is an iterative algorithm that uses stochastic sampling for a distribution of the optimization variables (parameters $K$ here) described by a mean and variance.
        As CMA is iterated, a cost function is used to evaluate random samples of $K$ and update the distribution so that it is centered around `better' samples. 
        
        We choose CMA because it is more robust to local minima, when compared to gradient based methods.
        This is because CMA is stochastic and, thus, considers a wider range of sample states.

    \subsection{Evaluating Performance}
    \label{Section: Cost Function}
    
        We formulate our task as an episodic reinforcement problem. 
        In our case, an episode refers to simulating RAMone with a specific set of parameters $K$ for a fixed time $t_{sim} {\,=\,} 7s$. 
        An episode ends early if RAMone falls.
        At the end of an episode, the performance of the set of parameters $K$ is evaluated with the following cost function: 
        
        \begin{equation} \label{Eqn: Cost Function}
            Cost =
            \begin{cases}
              100 + 20 \cdot \Delta t_{remaining}, & \text{robot falls} \\
              30 \cdot CoT + 1000 \cdot \left(\Delta \dot{x}_{des}\right)^2, & \text{otherwise} \\
            \end{cases}
        \end{equation}
        
        \noindent
        where $\Delta t_{remaining}$ is the amount of time between the fall and $t_{sim}$, $CoT$ is the cost of transport (as calculated in \cite{smit2017energetic}), and $\Delta \dot{x}_{des}$ is the difference between desired and actual speed of RAMone (average horizontal velocity of the main body). For the actual speed, we average over the last six steps.
        
        The constants in the cost function were heuristically chosen to satisfy three criteria: 1)  falling is always penalized more than walking; 2) falling earlier is penalized more than falling later; and 3) $CoT$ and $\Delta \dot{x}_{des}$ have approximately the same weighted importance in the cost function.
        
    \subsection{Initialization of CMA}
    
        We compute optimal walking parameters $K$ sequentially, for a range of different speeds.
        For each speed, we use CMA and the cost function \eqref{Eqn: Cost Function}; we initialize the CMA sample distribution using previously found optimal parameters $K$ for an adjacent speed.
        For the speed of $\unitfrac[0.4]{m}{s}$ at the first iteration, CMA is initialized using hand-tuned parameters. 
    
\section{Results and Discussion}

    With the approach described, we found control gain parameters $K$ that produced stable walking of RAMone in simulation for speeds between $\unitfrac[0.1]{m}{s}$ and $\unitfrac[1.0]{m}{s}$.
    In contrast, when using hand-tuning, we were only able to stabilize walking at speeds between $\unitfrac[0.4]{m}{s}$ and $\unitfrac[1.0]{m}{s}$. 
    
\begin{figure} [t]
        \centering
        \includegraphics[width=.45\textwidth]{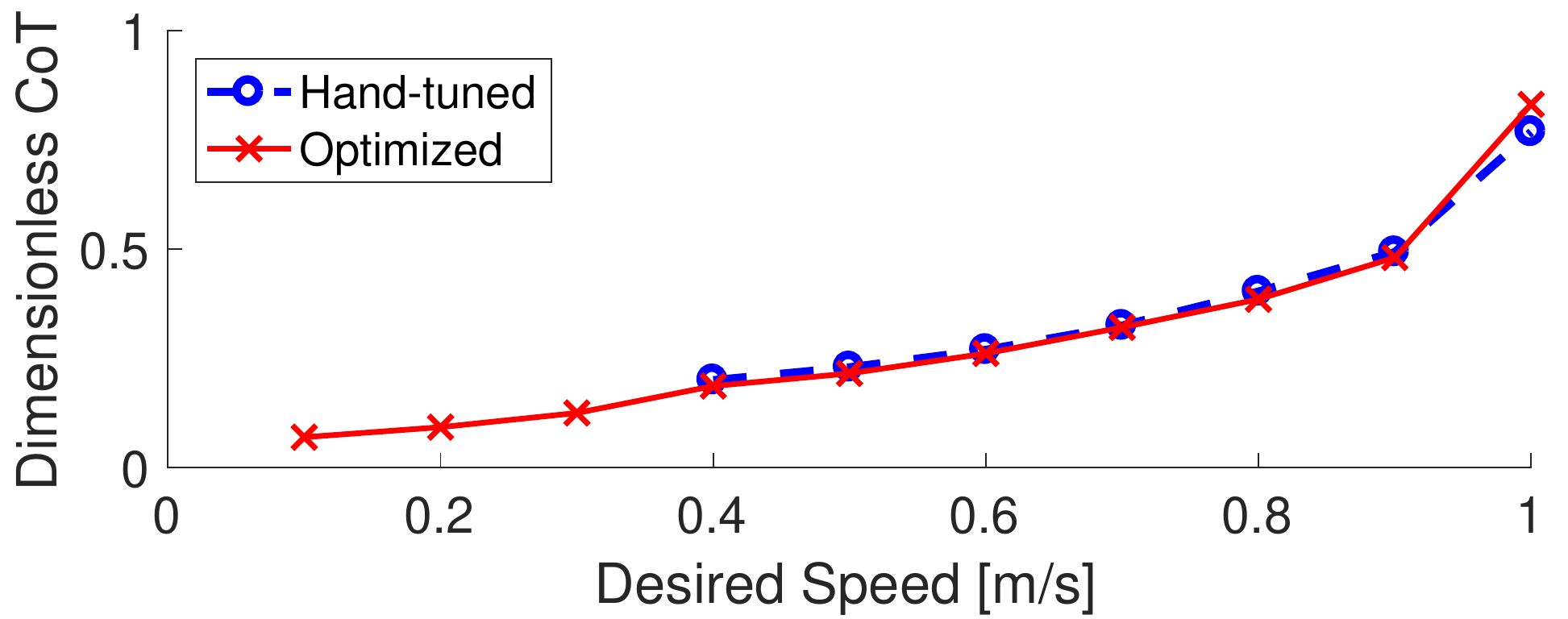}
        \caption{
            The cost of transport (CoT) across different speeds for control parameters obtained through hand-tuning (blue) and optimization (red). The CoT is similar in both cases, however optimization was able to find parameters for stable walking over a larger range of speeds.
        }
        \label{Fig: Compare CoT}
    \end{figure}

\begin{figure} [t]
        \centering
        \includegraphics[width=.45\textwidth]{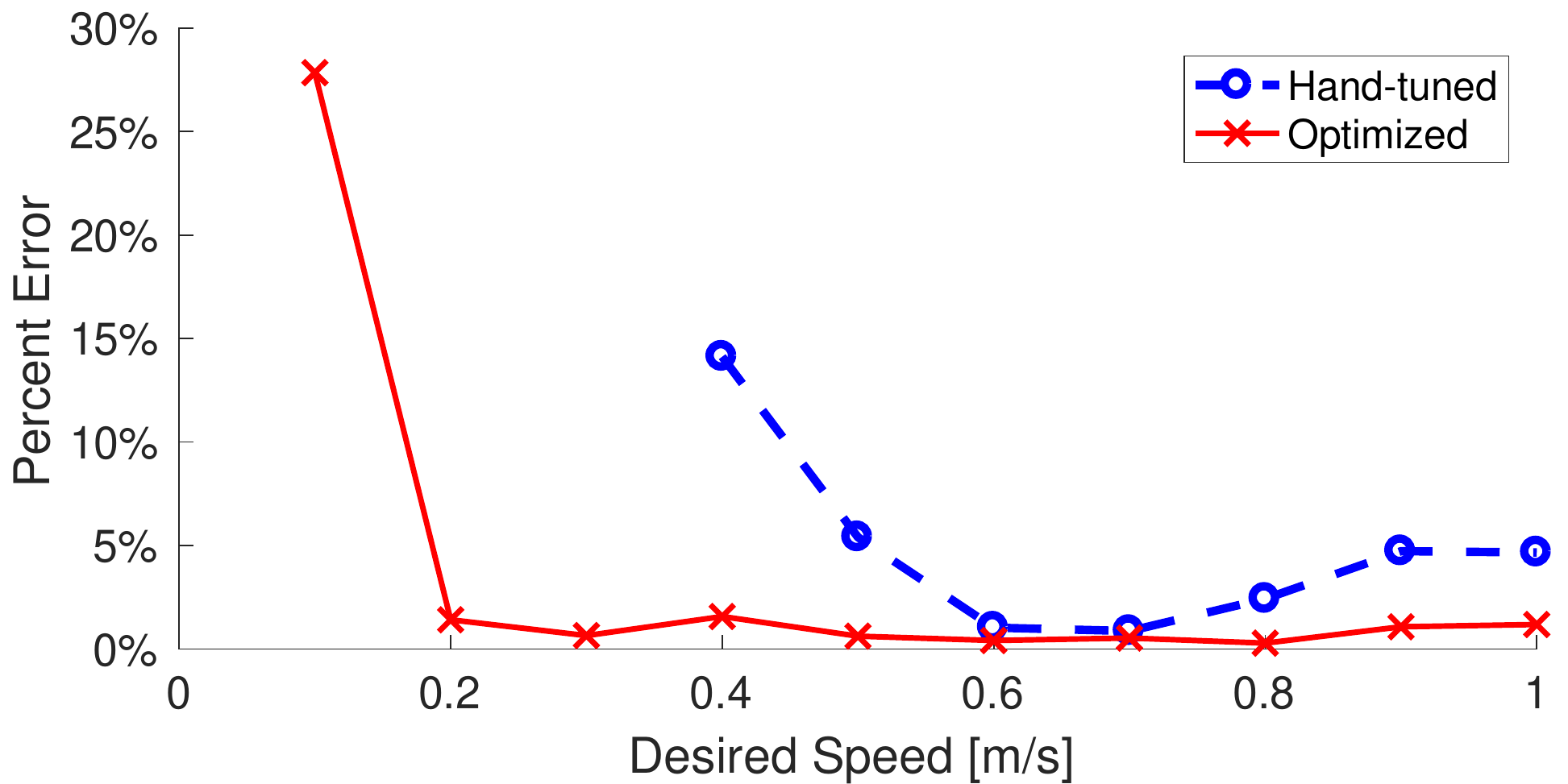}
        \caption{
            The percent error of RAMone's speed in simulation against the desired speed using control parameters obtained through hand-tuning (blue) and through optimization (red). Optimized parameters track the desired speed more accurately than hand-tuned parameters across all speeds. 
        }
        \label{Fig: Compare Speed Error}
    \end{figure}

    We found that the control parameters obtained through CMA yielded similar CoT, compared to the hand-tuned parameters, as shown in Fig.~\ref{Fig: Compare CoT}.
    There are two possible explanations for this: 
    1) The cost of transport does not depend strongly on the chosen parameters;
    and 2) our optimizer is getting caught in local minima and is thus not finding a more optimal solution.
    At the same time, the CMA-optimized controller performed better at desired speed tracking, compared to the hand-tuned controller, see Fig.~\ref{Fig: Compare Speed Error}.
    
    To continue this line of work, we plan to use the parameters found through optimization to achieve stable walking on hardware. 
    We also intend to modify the described method for application on hardware.
    
\section*{Acknowledgements}

    This material is based upon work supported by the National Science Foundation Graduate Research Fellowship under Grant No. DGE 1256260. Any opinion, findings, and conclusions or recommendations expressed in this material are those of the author(s) and do not necessarily reflect the views of the National Science Foundation.

\bibliography{references}
\bibliographystyle{plain}

\end{document}